\documentclass{article}
\usepackage[preprint]{neurips_2026}

\usepackage[utf8]{inputenc}
\usepackage[T1]{fontenc}
\usepackage{hyperref}
\usepackage{url}
\usepackage{booktabs}
\usepackage{amsmath,amsfonts,amsthm,bm}
\usepackage{nicefrac}
\usepackage{microtype}
\usepackage[table]{xcolor}
\usepackage{graphicx}
\usepackage[export]{adjustbox}
\usepackage{subcaption}
\usepackage{multirow}
\usepackage{enumitem}
\usepackage{xspace}
\usepackage{pifont}
\usepackage{float}

\usepackage{amsmath,amsfonts,amsthm, bm}

\def\eqref#1{equation~\ref{#1}}
\def\1{\bm{1}}

\def\mW{{\bm{W}}}
\def\mX{{\bm{X}}}

\DeclareMathAlphabet{\mathsfit}{\encodingdefault}{\sfdefault}{m}{sl}
\SetMathAlphabet{\mathsfit}{bold}{\encodingdefault}{\sfdefault}{bx}{n}

\def\sR{{\mathbb{R}}}

\newcommand{\abs}[1]{\vert#1\vert}

\ifx\proposition\undefined
\newtheorem{proposition}{Proposition}[section]
\fi

\newcolumntype{L}[1]{>{\raggedright\let\newline\\\arraybackslash\hspace{0pt}}m{#1}}
\newcolumntype{C}[1]{>{\centering\let\newline\\\arraybackslash\hspace{0pt}}m{#1}}
\newcolumntype{R}[1]{>{\raggedleft\let\newline\\\arraybackslash\hspace{0pt}}m{#1}}

\newcommand{\sect}[1]{Section~\ref{sect:#1}}
\newcommand{\app}[1]{Appendix~\ref{app:#1}}

\newcommand{\eqn}[1]{Equation~\ref{eq:#1}}

\newcommand{\fig}[1]{Figure~\ref{fig:#1}}

\newcommand{\tbl}[1]{Table~\ref{tab:#1}}

\newcommand{\lblfig}[1]{\label{fig:#1}}
\newcommand{\lblsect}[1]{\label{sect:#1}}
\newcommand{\lblapp}[1]{\label{app:#1}}
\newcommand{\lbleqn}[1]{\label{eq:#1}}
\newcommand{\lbltbl}[1]{\label{tab:#1}}

\newcommand{\ignorethis}[1]{}

\makeatletter
\DeclareRobustCommand\onedot{\futurelet\@let@token\@onedot}
\def\@onedot{\ifx\@let@token.\else.\null\fi\xspace}

\makeatother

\definecolor{citecolor}{rgb}{34,139,34}
\definecolor{mydarkblue}{rgb}{0,0.08,1}
\definecolor{mydarkgreen}{rgb}{0.02,0.6,0.02}
\definecolor{mydarkred}{rgb}{0.8,0.02,0.02}
\definecolor{mydarkorange}{rgb}{0.40,0.2,0.02}
\definecolor{mypurple}{RGB}{114, 92, 173}
\definecolor{myred}{RGB}{138, 0, 0}
\definecolor{mygold}{RGB}{234, 166, 77}
\definecolor{mydarkgray}{rgb}{0.66,0.66,0.66}
\definecolor{mitblue}{rgb}{0.88,0.95,0.96}
\definecolor{myblue}{RGB}{13, 94, 166}

\newcommand{\cmark}{\textcolor{green}{\ding{51}}}
\newcommand{\xmark}{\textcolor{red}{\ding{55}}}

\newif\ifdraft
\draftfalse
\ifdraft
    \newcommand{\Ali}[1]{{\color{myred}{\bf AK: #1}}}
    
    \newcommand{\Yann}[1]{{\color{myblue}{\bf YB: #1}}}
    
    \newcommand{\Sophie}[1]{{\color{mygold}{\bf SS: #1}}}
    
    \newcommand{\Mathieu}[1]{{\color{mydarkgreen}{\bf MS: #1}}}
    
    \newenvironment{paraphrase}
          {\color{mydarkgreen}\noindent\textbf{PARAPHRASING:}\ }
          {}
    \newenvironment{rewrite}
          {\color{mypurple}\noindent\textbf{MODIFY/IMPROVE:}\ }
          {}
\else
    \newcommand{\Ali}[1]{}
    
    \newcommand{\Yann}[1]{}
    
    \newcommand{\Sophie}[1]{}
    
    \newcommand{\Mathieu}[1]{}

\fi

\title{KroQuant: Kronecker-Structured Block Transforms for Efficient Post-Training Quantization of Diffusion Transformers}

\author{%
  Yann Bouquet\thanks{EPFL, Lausanne, Switzerland.\ \texttt{firstname.lastname@epfl.ch}}\kern0.55em\footnotemark[2]
  \quad
  Alireza Khodamoradi\thanks{Advanced Micro Devices, Inc., Longmont, CO, USA.\ \texttt{firstname.lastname@amd.com}}
  \quad
  Kristof Denolf\footnotemark[2]
  \quad
  Mathieu Salzmann\footnotemark[1]
}

\begin{document}
\maketitle

\begin{abstract}
Post-training quantization (PTQ) of diffusion transformers (DiTs) to W4A4 severely degrades output quality, because activations entering each linear layer contain outliers that 4-bit formats cannot represent. The standard fix applies an invertible linear transform to the activations and its inverse to the weights before quantizing both. Normalization layers between blocks force this transform to run online at every denoising step, making its inference computation cost the binding design constraint. Existing options trade quantization quality for inference cost: per-channel scaling (SmoothQuant) is computationally cheap but impacts the magnitude of the channels, which can harm quantization accuracy; fixed Hadamard transforms yield better quantization accuracy but require large block sizes that incur a high online cost; learned full-$d$ invertible transforms calibrate best but entail a prohibitive dense $d \times d$ matrix multiplication (GEMM) per layer per step.

We propose KroQuant, a PTQ method that applies a learned Kronecker-structured invertible transform to each 32-element block of the activation, storing less than half the parameters of per-channel scaling. The block-local structure runs as small tensor-core GEMMs, and on an MI350 GPU the KroQuant quantizer kernel is up to $14\%$ faster than the SmoothQuant kernel. Offline LoRaQ weight calibration then absorbs the residual per-weight quantization error. On PixArt-$\Sigma$, SANA, and FLUX.1-schnell at W4A4 (MXFP4e2), KroQuant produces outputs closer to the FP reference than SVDQuant and LoRaQ on MJHQ-30K and SDCI, while preserving or improving image quality.
\end{abstract}

\section{Introduction}\lblsect{intro}

\begin{figure}[t]
    \centering
    \begin{minipage}[c]{0.35\textwidth}
        \captionof{figure}{Pipeline comparison for W4A4 quantization of a linear layer in a DiT block. (\subref{fig:pipeline_svdquant})~SVDQuant splits the weight into a quantized residual and an SVD-truncated FP16 low-rank branch. (\subref{fig:pipeline_loraq})~LoRaQ replaces the SVD truncation by a data-free low-rank branch that can itself be quantized below 16 bits. (\subref{fig:pipeline_kroquant})~KroQuant additionally inserts a learned block-diagonal activation transform $T$ before quantization, complementing LoRaQ's weight-side correction. $P$ and $P_{SVD}$ are the 4-bit residual matrix defined as a function of the low rank components.}
        \lblfig{pipelines}
    \end{minipage}\hspace{1em}%
    \begin{minipage}[c]{0.6\textwidth}
        \centering
        \begin{subfigure}[t]{\linewidth}
            \centering
            \includegraphics[width=\linewidth]{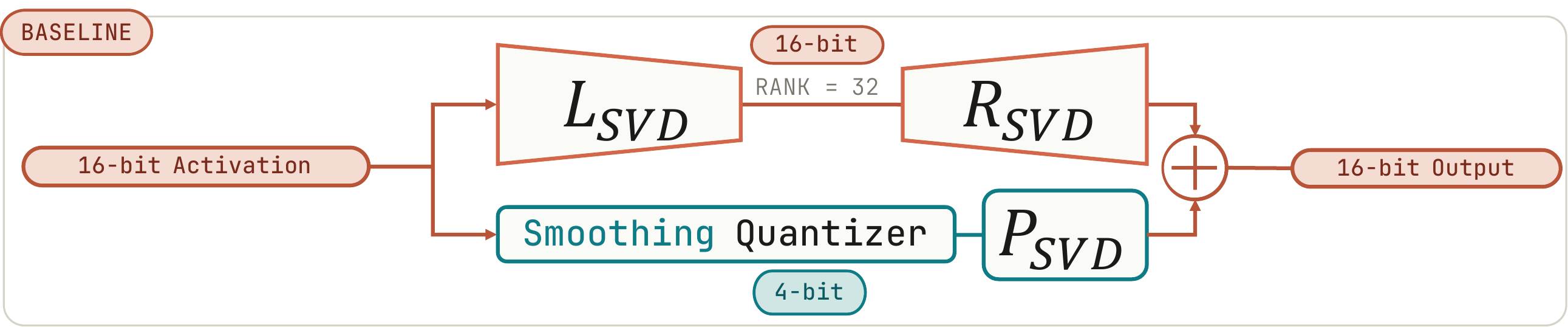}
            \caption{SVDQuant.}
            \lblfig{pipeline_svdquant}
        \end{subfigure}

        \begin{subfigure}[t]{\linewidth}
            \centering
            \includegraphics[width=\linewidth]{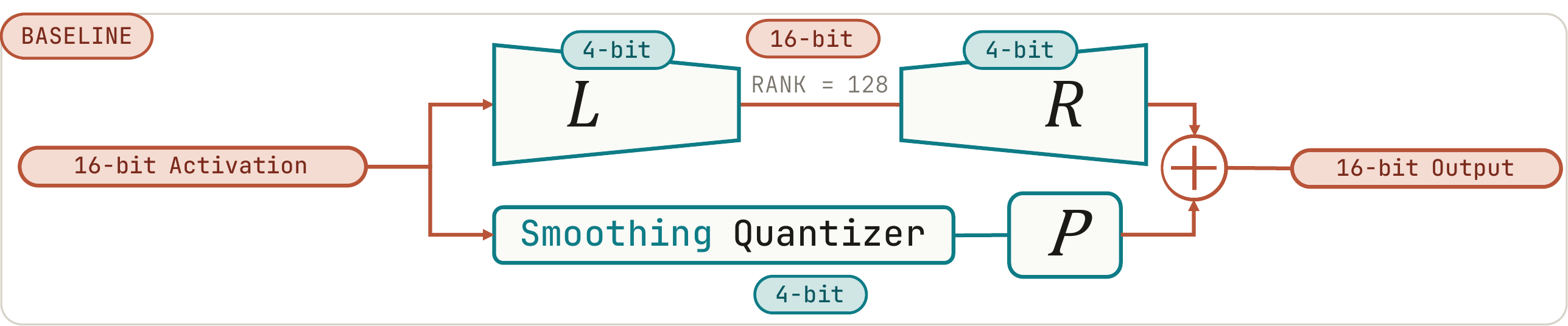}
            \caption{LoRaQ.}
            \lblfig{pipeline_loraq}
        \end{subfigure}

        \begin{subfigure}[t]{\linewidth}
            \centering
            \includegraphics[width=\linewidth]{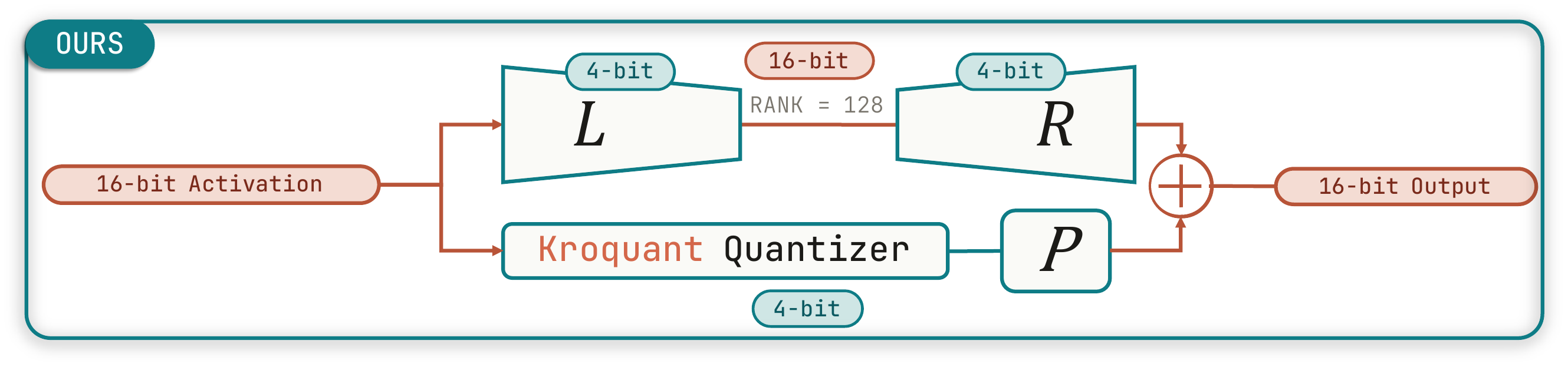}
            \caption{KroQuant.}
            \lblfig{pipeline_kroquant}
        \end{subfigure}
    \end{minipage}
\vspace{-15pt}
\end{figure}

Post-training quantization (PTQ) to 4-bit precision is the natural way to deploy billion-parameter diffusion transformers (DiTs), and is supported natively in current accelerators via the OCP Microscaling standard~\citep{ocp}. The bottleneck is activation outliers~\citep{xiao2023smoothquant}: a small subset of channels carries magnitudes orders of magnitude larger than the rest, so a $4$-bit blockwise quantizer wastes most of its dynamic range on those channels and clips everything else. Consider a linear layer $O = AW^\top$ with $A \in \mathbb{R}^{m\times d}$, $W \in \mathbb{R}^{o\times d}$, and a $b$-bit quantizer $Q_b$. The main solution in both LLM and DiT quantizers is to insert an invertible activation transform $T \in \mathrm{GL}(d)$ that redistributes channel magnitudes before quantization, acting as
\begin{equation}
  A \mapsto AT, \qquad W \mapsto WT^{-T}, \qquad AT \cdot (WT^{-\top})^\top = AT \cdot T^{-1}W^\top = AW^\top,
\end{equation}
which leaves the linear map invariant while reducing the per-layer quantization error
\begin{equation}
  \mathcal{E}(T) = Q_b(AT)\, Q_b(WT^{-\top})^\top - AW^\top.
\end{equation}
The compensating $T^{-1}$ is always folded into the layer's stored weights; the \emph{absorption trick} further folds $T$ into the previous layer's output, so that the activation arriving at the current layer is already $T$-transformed and the runtime transform is free. As such, existing methods differ only in the structure of $T$: diagonal per-channel scaling~\citep{xiao2023smoothquant, zhao2025viditq, wu2024ptq4dit}, dense rotations~\citep{quarot, quipsharp, liu2025spinquant}, or general invertible matrices~\citep{flatquant}.

While generally applicable to LLMs, the absorption trick fails in DiTs, especially between transformer blocks. This is because folding $T$ into the previous layer's output requires the normalization between blocks to commute with $T$:
\begin{equation}
  \mathrm{Norm}(xT)\, T^{-1} = \mathrm{Norm}(x).
\end{equation}
This is not satisfied by the AdaLN variant sitting between every DiT block (\app{commute}). Every transform a DiT quantizer applies therefore runs \textbf{online}, at every linear layer of every denoising step, and per-step cost becomes the binding design constraint. While the per-step cost is affordable with per-channel scaling transforms that entail diagonal matrices, this strategy offers only limited expressivity and thus quantization accuracy. This trade-off is taken by recent weight-splitting PTQ pipelines such as SVDQuant~\citep{li2025svdquant} and LoRaQ~\citep{loraq}, which decompose each weight into a low-bit residual and a low-rank correction branch.

By contrast, existing non-diagonal transforms yield better quantization but fall short under the online constraint. A Hadamard transform is a fixed orthogonal and symmetric matrix with normalized $\pm 1$ entries that mixes channels via add/subtract butterflies, spreading outlier magnitudes uniformly across the block~\citep{quarot}. At block size 32, it is cheap (a single MXFP4-aligned kernel pass) but suppresses outliers only within each 32-channel sub-vector, and eventually ends up underperforming per-channel scaling~\citep{xiao2023smoothquant}. Learned $d \times d$ invertible matrices~\citep{liu2025spinquant,flatquant} are expressive but store $O(d^2)$ parameters per layer and require a dense $m \times d \times d$ matrix multiplication every step, infeasible at inference. Altogether, there is therefore a need for a transform that is \textbf{learned} and \textbf{$32 \times 32$ block-diagonal}; that is, more expressive than a fixed Hadamard at the same block size, yet online-cheap and compatible with an MX quantizer.

To address this need, we propose \textbf{KroQuant}, a learned $T$ that is block-diagonal with $32 \times 32$ blocks and a compact parametrization that uses fewer scalars per layer than per-channel scaling, contains the Hadamard as a special case used as initialization, and applies as a single $32 \times 32$ tensor-core GEMM per block at inference; the construction is detailed in~\sect{method-2}. We show that KroQuant acts as a drop-in replacement for SmoothQuant~\citep{xiao2023smoothquant} in the activation-transform stage of weight-splitting PTQ pipelines such as LoRaQ (\fig{pipelines}). On PixArt-$\Sigma$~\citep{chen2024pixart-Sigma}, SANA~\citep{xie2025sana}, and FLUX.1-schnell~\citep{flux1}, LoRaQ-with-KroQuant matches or beats LoRaQ-with-SmoothQuant~\citep{loraq,xiao2023smoothquant} and the SVDQuant reference~\citep{li2025svdquant} at W4A4, while the KroQuant quantizer kernel runs up to $14\%$ faster than the SmoothQuant kernel on an MI350 GPU.

Concretely, we contribute:
\begin{itemize}[leftmargin=*]
  \item \textbf{A learned block-diagonal activation transform compatible with online constraints.} The first non-diagonal online transform for DiT PTQ that is more expressive than a fixed block-Hadamard and cheaper at inference than full-$d$ rotations~\citep{liu2025spinquant} or invertibles~\citep{flatquant}.
  \item \textbf{A Kronecker-LU parametrization of $\mathrm{GL}(32)$ blocks.} Five $2 \times 2$ unit-determinant LU factors yield a stable, $68\times$ smaller parameter footprint than dense $\mathrm{GL}(32)$ blocks, with the Hadamard as a natural initialization.
  \item \textbf{A fused W4A4 kernel.} Each block applies as a single tensor-core GEMM; on MI350 the KroQuant quantizer kernel is up to $14\%$ faster than the SmoothQuant baseline kernel.
\end{itemize}

\section{Related Work}\lblsect{related_work}

\paragraph{PTQ for diffusion transformers.}
DiT quantization has progressed from 8-bit schemes~\citep{shang2023ptqdm} to sub-8-bit methods that adapt SmoothQuant~\citep{xiao2023smoothquant} to the timestep-conditioned activation distributions induced by AdaLN. PTQ4DiT~\citep{wu2024ptq4dit} and ViDiT-Q~\citep{zhao2025viditq} apply per-channel diagonal scales with timestep-dependent calibration; Q-DiT~\citep{chen2024qditaccurateposttrainingquantization} and MixDQ~\citep{zhao2024mixdq} push further with mixed-precision and sensitivity-driven allocations. All keep the activation transform $T$ diagonal, and all degrade at W4A4, where the residual error is dominated by inter-channel correlations that no per-channel rescaling can remove.

\paragraph{The design space of activation transforms.}
PTQ methods that go beyond per-channel scaling differ in the sparsity pattern of the invertible transform $T \in \mathrm{GL}(d)$. SmoothQuant's diagonal $T$ rescales channels independently; QuaRot~\citep{quarot} and QuIP\#~\citep{quipsharp} use fixed dense Hadamard rotations, whose outlier-suppression strength grows with block size but whose full-$d$ butterfly does not map onto tensor cores; SpinQuant~\citep{liu2025spinquant} and FlatQuant~\citep{flatquant} learn a dense $T$ on the Stiefel manifold or in $\mathrm{GL}(d)$ respectively, at the cost of a $d \times d$ parameter matrix and a dense GEMM per layer. In LLMs all of these absorb offline into adjacent weights through RMSNorm; in DiTs they cannot, as discussed in~\sect{intro}, so per-step cost becomes load-bearing. KroQuant sits between SmoothQuant and FlatQuant: $T$ is block-diagonal with $32 \times 32$ blocks, each more expressive than a fixed Hadamard at the same block size, and applied online as parallel tensor-core GEMMs with fewer parameters per layer than SmoothQuant's diagonal scales.

\paragraph{Low-rank weight splitting and the role of the activation transform.}
A complementary line of work splits the linear-layer weight $W$ into a low-rank branch and a 4-bit residual. SVDQuant~\citep{li2025svdquant} smooths outliers and SVD-decomposes the smoothed weight with data-dependent low-rank calibration; LoRaQ~\citep{loraq} instead fits the low-rank branch data-free to the residual quantization error, allowing the low-rank branch itself to be quantized below 16 bits. Both methods split the weight but quantize the activation as-is, so they require an online activation transform with no normalization to fuse against, and both default to a SmoothQuant-style diagonal $T$. KroQuant is a drop-in replacement for that transform: we adopt LoRaQ's weight correction unchanged and substitute its SmoothQuant front-end with our learned block-diagonal $T$, yielding the headline LoRaQ-with-KroQuant comparison against LoRaQ-with-SmoothQuant, with SVDQuant as an additional reference.

\section{Method}\lblsect{method}
\vspace{-5pt}

We work within the formal setup of~\sect{intro} and instantiate the blockwise quantizer $Q_b$ left abstract there as follows.

\subsection{Quantization setup}\lblsect{method-1}
We denote by $Q_b$ (\sect{intro}) the blockwise quantize-then-dequantize operator at bit width $b$, taking a real matrix and returning a matrix of the same shape: each row of the input is partitioned into non-overlapping blocks of size $n$; each block is encoded as a shared scale $s$ (the set of representable scales, e.g., FP8) and a $n$ vector $\mathbf{q}$ in the chosen $b$-bit format (FP4 when $b=4$); the dequantized blocks are reassembled into the output. 
%We define $Q_4$ as the MXFP4e2 format quantizer, which uses 4 bits per value and E8M0 for the shared scale for each 32-element block.

\paragraph{Goal.} We want to construct a transform $T \in \mathrm{GL}(d)$ that is block-diagonal,
\begin{equation}
  T = \mathrm{diag}(K^{(1)}, K^{(2)}, \ldots, K^{(\lceil d/n \rceil)}),
\end{equation}

where each diagonal block $K^{(j)} \in \mathrm{GL}(32)$, learnable (unlike a fixed Hadamard), and whose parametrization is less than half the size of per-channel scaling and more than $68\times$ smaller than a dense $\mathrm{GL}(32)$ block. We aim to leave the linear map invariant ($AT \cdot (WT^{-1})^\top = AW^\top$) while reducing $\|\mathcal{E}\|_F$.

\subsection{Kronecker-structured invertible block transforms}\lblsect{method-2}

This structure of $T$ couples features within each group of $n$ channels while leaving groups independent, matching the block structure of the MXFP4e2 quantizer $Q_4$.

\paragraph{Kronecker product construction.} We parametrize each $n \times n$ block as a Kronecker product of $\log_2 n$ small matrices:
\begin{equation}
  K_n = G_1 \otimes G_2 \otimes \cdots \otimes G_{\log_2 n}, \qquad G_k \in \mathbb{R}^{2 \times 2}, \; k = 1, \ldots, \log_2 n.
\end{equation}
For $n = 32$ this gives $\log_2 32 = 5$ factors. Since the Kronecker product of invertible matrices is invertible, and since $(G_1 \otimes \cdots \otimes G_5)^{-1} = G_1^{-1} \otimes \cdots \otimes G_5^{-1}$, inversion reduces to five independent $2 \times 2$ inversions.

\paragraph{Unit-determinant LU parametrization.} To prevent degenerate scaling solutions, where the transform collapses one operand toward zero while inflating the other, we enforce $|\det G_k| = 1$ for every factor. We parametrize each $2 \times 2$ matrix via an LU decomposition with fixed unit determinant:
\begin{equation}
  G_k = \underbrace{\begin{pmatrix} 1 & 0 \\ c_k & 1 \end{pmatrix}}_{L_k} \underbrace{\begin{pmatrix} a_k & b_k \\ 0 & -1/a_k \end{pmatrix}}_{U_k}, \qquad a_k \neq 0.
\end{equation}
One can verify that $\det G_k = a_k \cdot (-1/a_k) = -1$, so $|\det G_k| = 1$ is satisfied for any $a_k \neq 0$. The upper-triangular factor $U_k$ uses a negative reciprocal on the $(2,2)$ entry so that the parametrization includes orthogonal reflections and rotations as special cases. Each $G_k$ is governed by three real parameters $(a_k, b_k, c_k)$; with five factors per block this gives \textbf{15 parameters per $32 \times 32$ block}, compared to $32^2 = 1024$ for an unconstrained invertible matrix and 1 for a scalar channel-wise scale.

\paragraph{Hadamard initialization.} The normalized Hadamard matrix $H_n$ ($n = 2^k$) is a Kronecker product of $k$ copies of the $2\times2$ factor $\frac{1}{\sqrt{2}}\begin{pmatrix}1 & 1 \\ 1 & -1\end{pmatrix}$. It has determinant $-1$ and therefore satisfies $|\det H_n| = 1$. We initialize every $G_k$ as $G_k^{(0)} = \begin{pmatrix} \cos\!\left(\tfrac{\pi}{4}\right) & \sin\!\left(\tfrac{\pi}{4}\right) \\ \sin\!\left(\tfrac{\pi}{4}\right) & -\cos\!\left(\tfrac{\pi}{4}\right) \end{pmatrix}$ which is equivalent to initializing $T$ to a block-diagonal Hadamard rotation. This is known empirically as a good initialization strategy to reduce activation outliers and lower quantization error before any gradient-based optimization~\citep{quarot, liu2025spinquant}.

\subsection{Layer-wise transform optimization}\lblsect{method-3}

\paragraph{Objective.} Given a small calibration dataset providing $N$ activation samples $\{A^{(s)}\}_{s=1}^N$, we optimize $T$ independently for each linear layer by minimizing the output mean squared error:
\vspace{-5pt}
\begin{equation}
  \mathcal{L}_\text{out}(T) = \frac{1}{Nmo} \left\| Q_b(AT)\, Q_b(WT^{-T})^\top - AW^\top \right\|_F^2.
\end{equation}
\paragraph{Regularization.} Minimizing $\mathcal{L}_\text{out}$ alone can converge to solutions where one operand is well-quantized at the expense of the other --- for example, a transform that whitens the activations perfectly while leaving the weights in a highly non-uniform representation. To encourage both operands to be individually well-quantized, we add two auxiliary terms:
\begin{equation}
  \mathcal{L}_A(T) = \frac{1}{Nmd} \left\| Q_b(AT) - AT \right\|_F^2, \qquad
  \mathcal{L}_W(T) = \frac{1}{od} \left\| Q_b(WT^{-\top}) - WT^{-\top} \right\|_F^2.
\end{equation}
The total loss is
\begin{equation}
\lbleqn{objective}
  \mathcal{L}(T) = \mathcal{L}_\text{out}(T) + \lambda_A\, \mathcal{L}_A(T) + \lambda_W\, \mathcal{L}_W(T).
\end{equation}

\paragraph{Automatic scaling of $\lambda$.} The regularization weights $\lambda_A$ and $\lambda_W$ are set so that each auxiliary term matches the magnitude of $\mathcal{L}_\text{out}$ at initialization:
\begin{equation}
  \lambda_A = \frac{\mathcal{L}_\text{out}(T_0)}{\mathcal{L}_A(T_0)}, \qquad \lambda_W = \frac{\mathcal{L}_\text{out}(T_0)}{\mathcal{L}_W(T_0)},
\end{equation}
where $T_0$ is the Hadamard initialization. This normalization ensures that none of the three terms dominates early in training, preventing the regularizers from overriding the primary output-error signal during the initial phase of optimization, while still providing sufficient gradient signal to avoid degenerate solutions.
\vspace{-5pt}
\paragraph{Optimization.} We optimize $\mathcal{L}(T)$ with respect to the $15 \cdot \lceil d/n \rceil$ free parameters of $T$ using Adam. The quantizer $Q_b$ is non-differentiable; we use a straight-through estimator (STE) to propagate gradients through the quantization operations. Since the transforms of the different layers are independent, they can be optimized in parallel or sequentially with negligible overhead relative to the calibration forward passes. The use of an STE and the Kronecker composition of the optimizable parameters makes the optimization unstable at a block or model level which is the reason why we keep the optimization at a layer level.

\subsection{Final correction: offline LoRaQ weight calibration}\lblsect{method-4}
After optimizing the per-layer Kronecker transforms $\{T^{(\ell)}\}$ to minimize our objective, we apply the offline weight calibration procedure from LoRaQ~\citep{loraq} to the transformed weights $\widetilde{W}^{(\ell)} = W^{(\ell)} (T^{(\ell)})^{-\top}$.

~\citet{loraq} decomposes each transformed weight $\widetilde{W}^{(\ell)}$ into a $b$-bit quantized residual branch and an additive low-rank correction branch $L^{(\ell)} R^{(\ell)}$, with $L^{(\ell)} \in \mathbb{R}^{o \times r}$ and $R^{(\ell)} \in \mathbb{R}^{r \times d}$, and optimizes both branches data-free to carry most information of the 16-bit activation to the low-rank branch, reducing the  linear layer's overall quantization error further. The low-rank branch can itself be quantized below 16 bits without breaking the decomposition. At inference, the layer evaluates the quantized residual matrix multiplication and the mixed-precision low-rank correction in parallel and sums their outputs.

At inference time the computation for layer $\ell$ is
\begin{equation}
  \hat{O}^{(\ell)} = Q_b\!\left(A^{(\ell)} T^{(\ell)}\right) \times Q_b\!\left(\widetilde{W}^{(\ell)} - L^{(\ell)} R^{(\ell)}\right) + A^{(\ell)} T^{(\ell)}L^{(\ell)} R^{(\ell)},
\end{equation}
with $\hat{W}^{(\ell)}$ precomputed offline.

\subsection{Hardware-efficient transform application}\lblsect{method-5}

We implement the block-diagonal transform $T$ as two custom Triton kernels that apply each $32 \times 32$ block $K^{(j)}$ to the corresponding 32-element slice of every token. Both kernels tile the activation matrix $A \in \mathbb{R}^{m \times d}$ on a 2D grid of shape $(\lceil m / B_R \rceil,\; d/32)$, where $B_R \in \{16, 32, 64, 128\}$ is autotuned. Each thread block processes a tile of shape $(B_R, 32)$.

\subsubsection{Kernel variants}

\paragraph{K1 (precomputed dense).} Each $32 \times 32$ block $K^{(j)}$ is materialized and stored in HBM. The kernel loads the activation tile $A_\text{tile} \in \mathbb{R}^{B_R \times 32}$ and the corresponding $K^{(j)} \in \mathbb{R}^{32 \times 32}$, computes a single \texttt{tl.dot($A_\text{tile}$, $K_\text{block}$)} accumulating in fp32, and stores the result after casting to the input dtype. Storage cost: 1024 values (2 KB) per column block; 256 KB total for $d = 4096$.
\vspace{-10pt}
\paragraph{K2 (on-the-fly Kronecker construction).} Instead of loading a dense matrix, K2 reconstructs $K^{(j)}$ from the 15 scalar parameters of the five $2 \times 2$ factors $G_1, \ldots, G_5$ (\sect{method-2}). Since $32 = 2^5$, every row and column index of $K^{(j)}$ is a $5$-bit integer, and the Kronecker product entry-wise formula reads
\begin{equation}
  K[r,c] = \prod_{k=1}^{5} G_k\!\bigl[r_{5-k},\, c_{5-k}\bigr],
\end{equation}
where $r_p$ and $c_p$ denote the $p$-th bit of $r$ and $c$ respectively. The kernel applies this formula as a streaming product over the five factors, materializing $K$ factor-by-factor in registers via bit-indexed scalar lookups, leaving the register budget to fuse the downstream MXFP4e2 quantizer in the same kernel. Pseudocode is in~\app{k2-kernel}. Storage cost: 15 scalar parameters per column block, a $\approx68\times$ compression over K1.

Both kernels satisfy the Triton \texttt{tl.dot} constraint $M, N, K \geq 16$ for all $B_R \geq 16$, since the GEMM shape is $(B_R, 32) \times (32, 32) \to (B_R, 32)$.

K2 stores 15 scalar parameters per block ($68\times$ compression over K1's 1024 values) at a throughput cost of only 5--15\% relative to K1 (see~\app{appendix-kernel}). For models with large hidden dimensions where many column blocks share the same Kronecker factors, K2's parameter compression reduces HBM pressure on the transform-matrix read path. For inference regimes where transform storage is not a bottleneck, K1 offers higher absolute throughput.

%\subsubsection{K2 vs.\ FWHT: structural bandwidth advantage}
%FWHT requires $\log_2 d$ sequential passes over the data (12 passes for $d = 4096$), yielding arithmetic intensity $\text{AI} \approx 0.25$ FLOP/byte --- deeply bandwidth-wasteful. K2 is single-pass with $\text{AI} \approx 16$ FLOP/byte. The speedup is structural, not implementational: it grows with $d$ because FWHT pass count scales as $\log_2 d$ while K2 scales linearly with $d/32$.

\section{Experiments}\lblsect{experiments}

\begin{table}[b]
    \vspace{-10pt}
  \caption{(MXFP4 E2M1, W4A4) on MJHQ-30K and SDCI (5k samples each). LR decomposition rank is 128, MXFP4e2 dataformat for the LoRaQ backbone. Rank 32, 16-bit dataformat for the LR decomposition in SVDQuant. Best quantized result per model and dataset in \textbf{bold}.}
  \lbltbl{main}
  \centering
  \resizebox{\textwidth}{!}{%
  \begin{tabular}{lllcccccccc}
    \toprule
    & & & \multicolumn{4}{c}{MJHQ-30K} & \multicolumn{4}{c}{SDCI} \\
    \cmidrule(lr){4-7}\cmidrule(lr){8-11}
    Model & Quantizer & LR Decomposition & FID $\downarrow$ & IR $\uparrow$ & LPIPS $\downarrow$ & PSNR $\uparrow$ & FID $\downarrow$ & IR $\uparrow$ & LPIPS $\downarrow$ & PSNR $\uparrow$ \\
    \midrule
    \multirow{4}{*}{PixArt-$\Sigma$}
      & \multicolumn{2}{l}{FP16 (reference)}              & 16.6          & 0.948          & ---            & ---           & 24.8          & 0.966          & ---            & ---           \\
      & Smoothing                 & SVDQuant              & 19.3          & 0.876          & 0.433          & 15.2          & 23.3          & 0.904          & 0.476          & 13.9          \\
      & Smoothing                 & LoRaQ                 & 17.5          & 0.952          & 0.381          & 16.4          & 23.7          & 0.970          & 0.399          & 15.0          \\
      & \textbf{KroQuant (Ours)}  & LoRaQ                 & \textbf{16.9} & \textbf{0.989} & \textbf{0.347} & \textbf{16.5} & \textbf{21.9} & \textbf{1.01}  & \textbf{0.378} & \textbf{15.2} \\
    \midrule
    \multirow{4}{*}{SANA}
      & \multicolumn{2}{l}{BF16 (reference)}              & 16.2          & 1.09           & ---            & ---           & 22.4          & 1.07           & ---            & ---           \\
      & Smoothing                 & SVDQuant              & \textbf{16.6} & 1.07           & 0.256          & 17.1          & 23.8          & 1.02           & 0.285          & 15.4          \\
      & Smoothing                 & LoRaQ                 & 17.4          & 1.09           & 0.229          & 17.7          & 24.1          & 1.03           & 0.259          & 15.9          \\
      & \textbf{KroQuant (Ours)}  & LoRaQ                 & \textbf{16.6} & \textbf{1.09}  & \textbf{0.206} & \textbf{18.4} & \textbf{23.6} & \textbf{1.04}  & \textbf{0.241} & \textbf{16.4} \\
    \midrule
    \multirow{4}{*}{FLUX.1-schnell}
      & \multicolumn{2}{l}{BF16 (reference)}              & 19.2          & 0.938          & ---            & ---           & 20.8          & 0.932          & ---            & ---           \\
      & Smoothing                 & SVDQuant              & 20.1          & \textbf{0.952} & 0.378          & 15.7          & 22.4          & 0.869          & 0.589          & 13.3          \\
      & Smoothing                 & LoRaQ                 & 19.9          & 0.921          & \textbf{0.331} & 16.7          & 22.0          & 0.961          & 0.343          & 15.8          \\
      & \textbf{KroQuant (Ours)}  & LoRaQ                 & \textbf{19.8} & 0.880          & 0.336          & \textbf{16.9} & \textbf{21.2} & \textbf{0.982} & \textbf{0.283} & \textbf{16.6} \\
    \bottomrule
  \end{tabular}}
\end{table}
\vspace{-5pt}
\subsection{Experimental setup}\lblsect{exp-1}

\paragraph{Models.} We evaluate on three diffusion transformers of increasing scale and architectural diversity:
\begin{itemize}[leftmargin=*]
  \item \textbf{PixArt-$\Sigma$} \citep{chen2024pixart-Sigma} with 0.6 billion parameters, a single-stream architecture for text-to-image generation.
  \item \textbf{SANA} \citep{xie2025sana} with 1.6 billion parameters, a single-stream architecture for text-to-image generation with 1x1 convolution layers in the FFN module.
  \item \textbf{FLUX.1-schnell} \citep{flux1} with 12 billion parameters, a double-stream architecture for text-to-image generation.
\end{itemize}

\paragraph{Calibration dataset.} Following the protocol of SVDQuant \citep{li2025svdquant} and LoRaQ \citep{loraq}, we use 128 text prompts drawn from the COCO 2017 validation captions~\citep{chen2015coco} as calibration inputs. For each prompt a single forward pass is executed per linear layer to collect activation statistics used in transform optimization and LoRaQ weight calibration.
\vspace{-10pt}
\paragraph{Evaluation datasets.} We report image quality on two different datasets, MJHQ-30K \citep{li2024mjhq} and SDCI~\citep{urbanek2024sdci}, each with 5,000 images randomly sampled  with the same fixed-seed protocol following the evaluation split used in prior work.
\vspace{-10pt}
\paragraph{Metrics.} Following existing benchmarks, we evaluate performance on two criteria. To measure similarity to the 16-bit baseline, we use Learned Perceptual Image Patch Similarity (LPIPS)~\citep{zhang2018perceptual} and Peak Signal-to-Noise Ratio (PSNR). To assess overall visual quality, we use Frechet Inception Distance (FID)~\citep{heusel2018ganstrainedtimescaleupdate}, Image Reward (IR)~\citep{xu2023imagereward}.
\vspace{-10pt}
\paragraph{Data format.} We use MXFP4 E2M1 format following the OCP Microscaling (MX) standard~\citep{ocp}, which is motivated by the fact that current hardware and software stacks, including \citet{AMD_CDNA4_ISA_2025} and emerging ML frameworks, natively support MX formats at this granularity, as also adopted by~\citet{loraq}. We refer to~\app{config} for a description of the blockwise quantizer.

We provide a full description of the experiments for reproducibility in~\app{config}.

\subsection{Main quantitative results}\lblsect{exp-2}

~\tbl{main} compares KroQuant against prior PTQ methods on PixArt-$\Sigma$, SANA, and FLUX.1-schnell across two benchmarks. The smoothing method used by SVDQuant and LoRaQ are the same as described in~\citet{li2025svdquant, loraq}. KroQuant matches or improves upon all baselines on PixArt-$\Sigma$ and SANA across both benchmarks. On MJHQ-30K, KroQuant reduces LPIPS by $0.034$ (relative $8.9\%$) over LoRaQ on PixArt-$\Sigma$ and by $0.023$ ($10.0\%$) on SANA, while closing the FID gap to the full-precision reference (16.9 vs.\ 16.6 on PixArt-$\Sigma$; matching 16.6 on SANA). The SDCI benchmark confirms these trends: KroQuant achieves the best FID, LPIPS, and PSNR on all three models, with particularly large LPIPS gains on FLUX.1-schnell ($0.283$ vs.\ $0.343$ for LoRaQ, a $17.5\%$ relative reduction).

On FLUX.1-schnell / MJHQ-30K, the picture is mixed: KroQuant achieves the best FID and PSNR but fails to improve over SVDQuant on IR and LoRaQ on LPIPS. We attribute this to the double-stream architecture of FLUX.1, where joint-attention blocks interleave text and image tokens with different activation distributions; a single per-layer Kronecker transform may underfit this heterogeneity. The consistent SDCI gains on FLUX.1 suggest the deficit is benchmark-dependent rather than systematic.
\vspace{-20pt}
\paragraph{Visual Results.}~\fig{kroquant_visual_comparison} shows a qualitative comparison of KroQuant against LoRaQ and SVDQuant on PixArt-$\Sigma$ / MJHQ-30K. This figure shows that KroQuant produces sharper images with fewer artifacts than the baselines, consistent with the improved quantitative metrics. We provide additional visual comparisons in~\app{vis}.

\begin{figure*}[t]
    \centering
    \caption{Visual comparison on PixArt-$\Sigma$ at MXFP4e2 W4A4. \textbf{Left three columns}: FP16 (reference), SVDQuant, KroQuant; KroQuant uses the LoRaQ backbone for offline weight calibration while SVDQuant uses its own SVD-based low-rank decomposition. \textbf{Right three columns}: FP16 (reference), LoRaQ leveraging the SmoothQuant quantizer, and KroQuant relying on the LoRaQ backbone.}
    \lblfig{kroquant_visual_comparison}
    \setlength{\tabcolsep}{2pt}
    \begin{tabular}{ccc@{\hspace{0.6em}}ccc}
        \toprule
        {\small \textbf{FP16}} & {\small \textbf{SVDQuant}} & {\small \textbf{KroQuant (Ours)}} & {\small \textbf{FP16}} & {\small \textbf{LoRaQ}} & {\small \textbf{KroQuant (Ours)}} \\
        \midrule

        % --- ROW 1: SVDQuant comparison sample_08 | LoRaQ comparison sample_04 ---
        \adjustbox{frame,width=0.15\textwidth}{
            \includegraphics{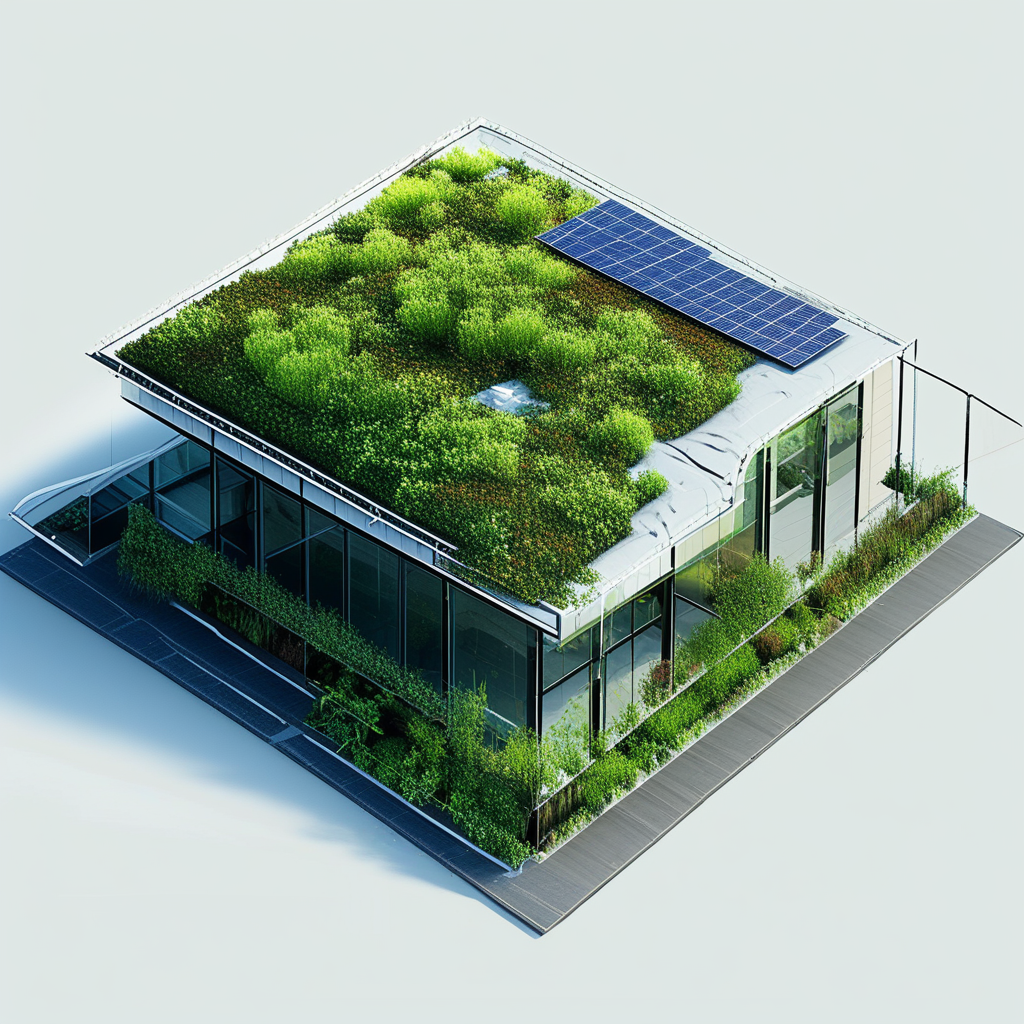}} &
        \adjustbox{frame,width=0.15\textwidth}{
            \includegraphics{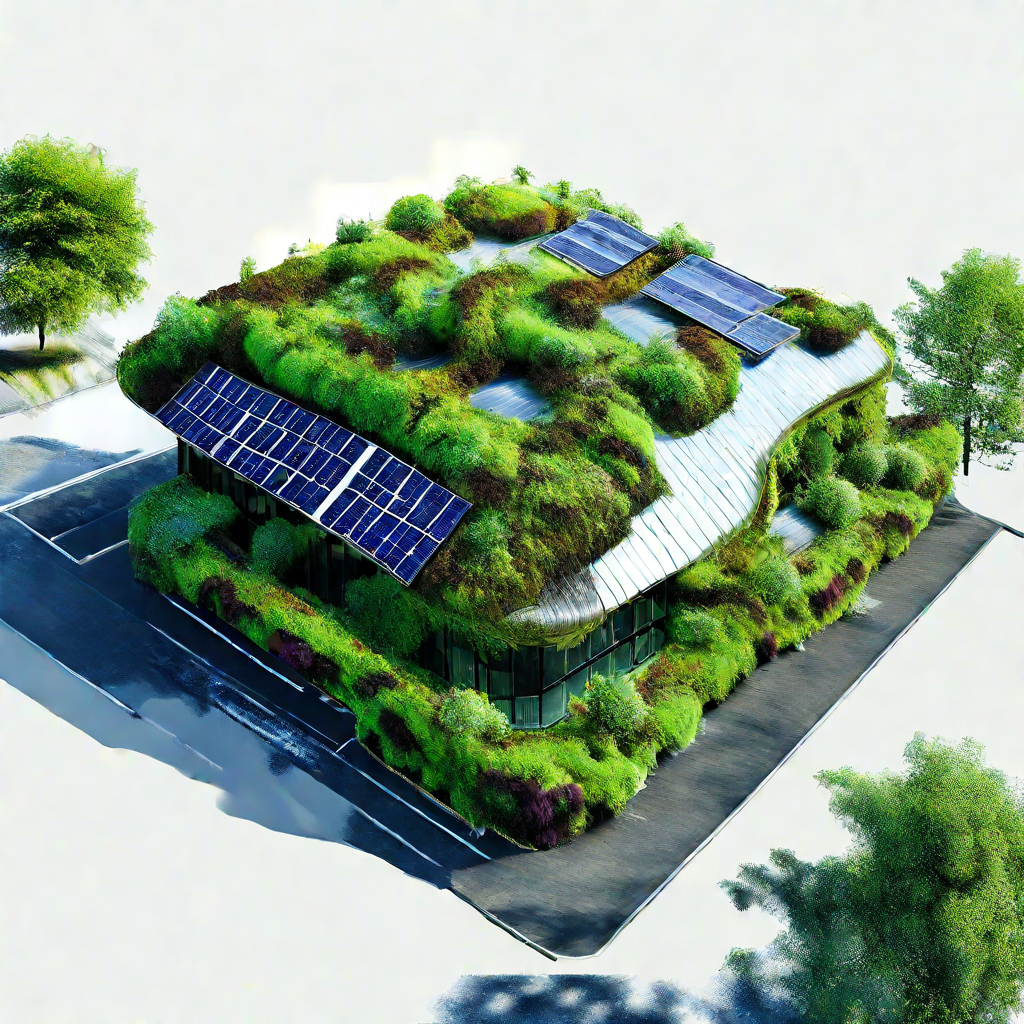}} &
        \adjustbox{frame,width=0.15\textwidth}{
            \includegraphics{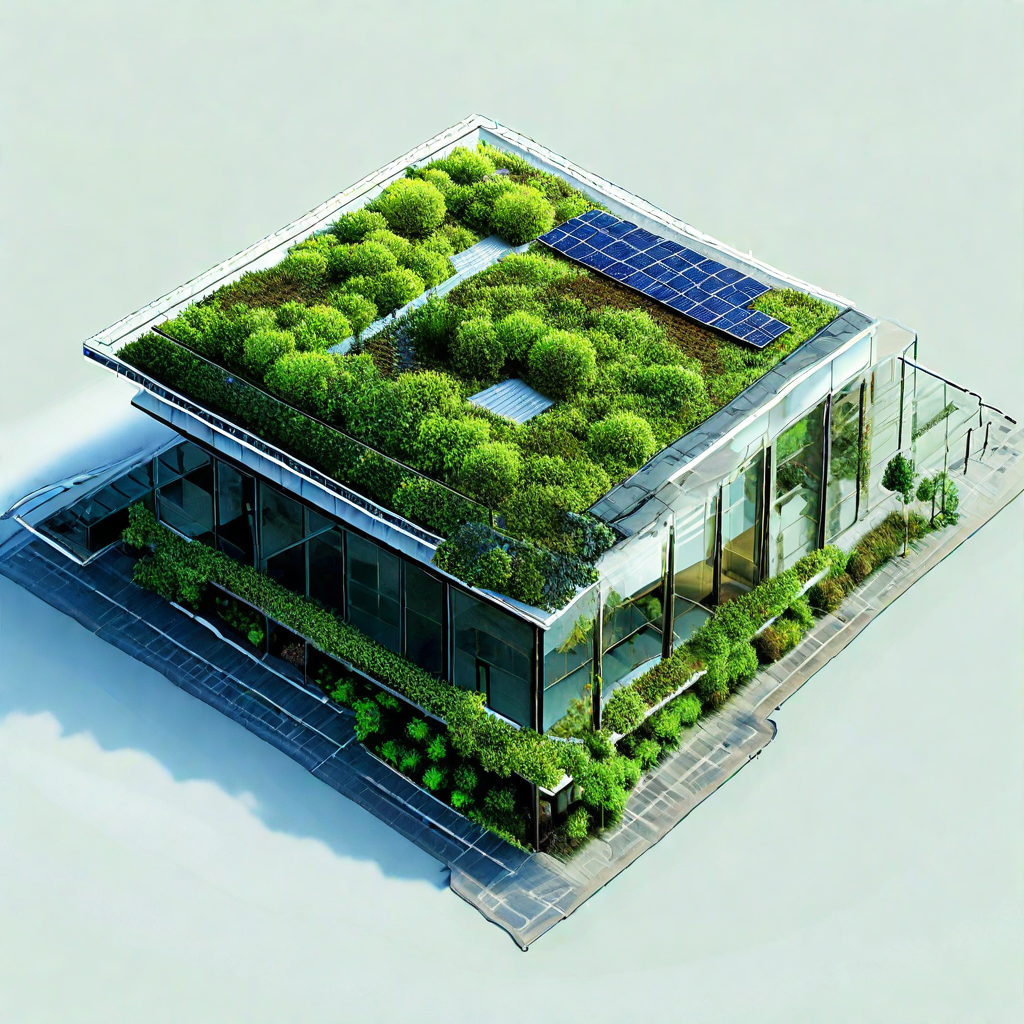}} &
        \adjustbox{frame,width=0.15\textwidth}{
            \includegraphics{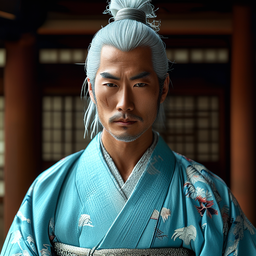}} &
        \adjustbox{frame,width=0.15\textwidth}{
            \includegraphics{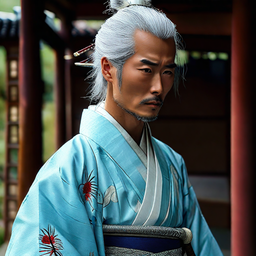}} &
        \adjustbox{frame,width=0.15\textwidth}{
            \includegraphics{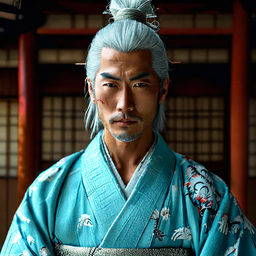}} \\

        % --- ROW 2: SVDQuant comparison sample_04 | LoRaQ comparison sample_02 ---
        \adjustbox{frame,width=0.15\textwidth}{
            \includegraphics{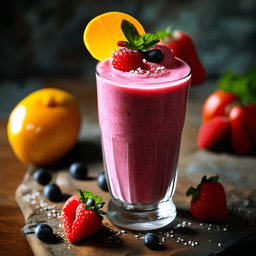}} &
        \adjustbox{frame,width=0.15\textwidth}{
            \includegraphics{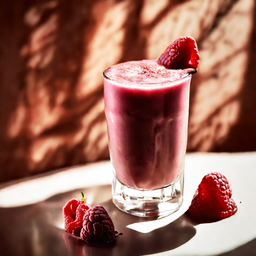}} &
        \adjustbox{frame,width=0.15\textwidth}{
            \includegraphics{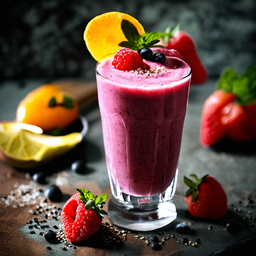}} &
        \adjustbox{frame,width=0.15\textwidth}{
            \includegraphics{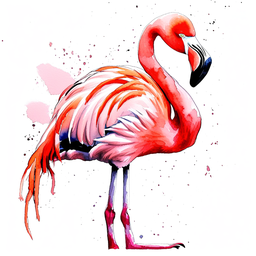}} &
        \adjustbox{frame,width=0.15\textwidth}{
            \includegraphics{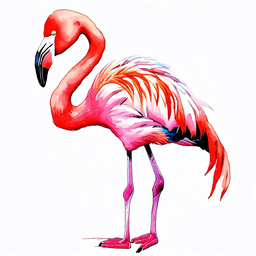}} &
        \adjustbox{frame,width=0.15\textwidth}{
            \includegraphics{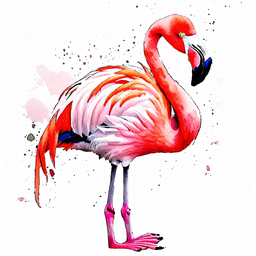}} \\

        \adjustbox{frame,width=0.15\textwidth}{
            \includegraphics{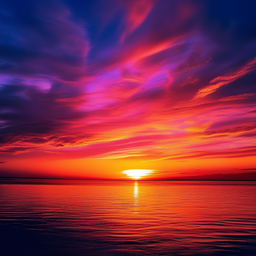}} &
        \adjustbox{frame,width=0.15\textwidth}{
            \includegraphics{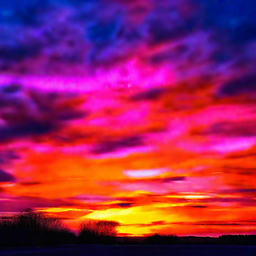}} &
        \adjustbox{frame,width=0.15\textwidth}{
            \includegraphics{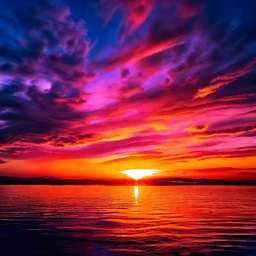}} &
        \adjustbox{frame,width=0.15\textwidth}{
            \includegraphics{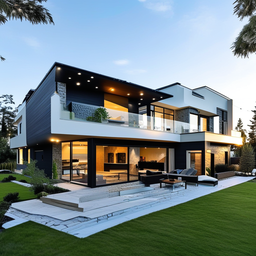}} &
        \adjustbox{frame,width=0.15\textwidth}{
            \includegraphics{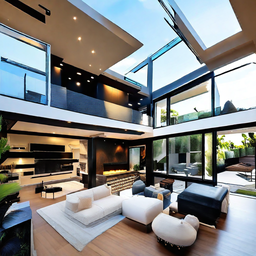}} &
        \adjustbox{frame,width=0.15\textwidth}{
            \includegraphics{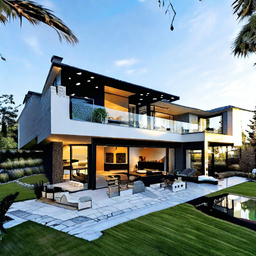}} \\
            
        \bottomrule
    \end{tabular}
\end{figure*}

\subsection{Ablation study}\lblsect{ablation}

\begin{table}[t]
\vspace{-15pt}
  \caption{Ablation: transform design choices on PixArt-$\Sigma$ / MJHQ-30K (5k samples). All quantized rows use MXFP4 E2M1, $b=32$, W4A4. ``--'' in the LR Decomposition column indicates no offline weight-correction stage.}
  \lbltbl{ablation}
  \centering
  \begin{tabular}{llccc}
    \toprule
    Quantizer & LR Decomposition & FID $\downarrow$ & LPIPS $\downarrow$ & PSNR $\uparrow$ \\
    \midrule
    Smoothing                          & --  & 130.60        & 0.712              & 13.0            \\
    \textbf{KroQuant (Ours)}           & --  & 41.8          & 0.516              & 14.8            \\
    \midrule
    Smoothing                          & LoRaQ   & 17.5          & 0.381              & 16.4            \\
    Hadamard ($32{\times}32$)          & LoRaQ   & 17.5          & 0.382              & 15.7            \\
    \textbf{KroQuant (Ours)}           & LoRaQ   & \textbf{16.9} & \textbf{0.347}     & \textbf{16.5}   \\
    \bottomrule
  \end{tabular}
\vspace{-15pt}
\end{table}

\tbl{ablation} ablates the transform family on PixArt-$\Sigma$ / MJHQ-30K. The upper group shows standalone transforms without LoRaQ; the lower group holds LoRaQ calibration fixed across all rows.

The standalone rows (without LoRaQ) isolate the transform contribution against the scaling method Smoothing: KroQuant alone reduces FID from 130.6 to 41.8, a large improvement but still far from full-precision quality. This confirms that LoRaQ weight calibration is essential for closing the remaining gap, while the transform provides a better starting point for weight calibration to operate on.

With LoRaQ, the block Hadamard-32 baseline matches per-channel smoothing on FID (both 17.5) but slightly degrades PSNR (15.7 vs.\ 16.4), indicating that a fixed Hadamard rotation does not uniformly reduce quantization error and can be mildly harmful for PSNR-sensitive layers. The gap between Block Hadamard-32 and Kronecker (Hadamard init), $\Delta$LPIPS $= 0.035$, $\Delta$PSNR $= 0.8$ dB, demonstrates that the 15 learnable parameters per block genuinely improve upon the fixed rotation, capturing per-layer numerical structure that a fixed orthogonal transform cannot adapt to.

\subsection{Inference latency}\lblsect{exp-4}

\begin{figure}[b]
  \centering
  \includegraphics[width=\linewidth]{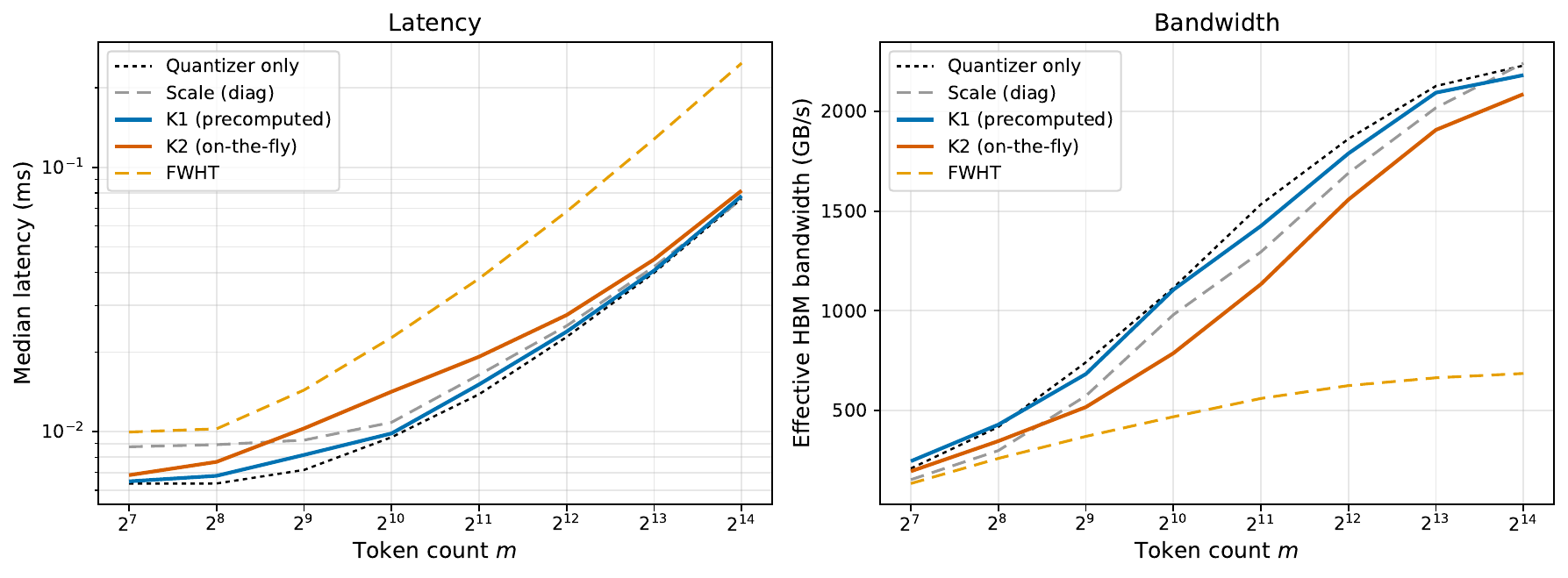}
  \caption{Median kernel latency and effective HBM bandwidth vs.\ token count $m$. $d=4096$ dimensions, evaluated on MI350 with FP16 and MXFP4e2 quantizer. K1 and K2 track Scale closely and is ${\sim}1.4\times$ faster than FWHT across all $m$.}
  \lblfig{latency}
\end{figure}

We measure the wall-clock cost of the KroQuant block-diagonal transform (\sect{method-5}) against per-channel scaling (\textbf{Scale}, a lower bound on per-layer arithmetic intensity) and the Fast Walsh--Hadamard Transform (\textbf{FWHT}~\citep{quarot, liu2025spinquant, quipsharp}).
\vspace{-10pt}
\paragraph{Setup.} All kernels run on an AMD Instinct MI350 in FP16, fuse a block-wise MXFP4e2 quantizer at their output, and are timed as medians over 500 repetitions after 100 warmup calls; reported latencies therefore reflect the full per-layer quantizer rather than the bare transform.

We benchmark two KroQuant variants: \textbf{K1} loads precomputed dense $32 \times 32$ Kronecker blocks from memory, while \textbf{K2} reconstructs each block on the fly from $15$ learned parameters ($68\times$ memory compression vs.\ K1). FWHT requires power-of-two $d$, which forces zero-padding (extra memory traffic) or truncation (information loss) on production shapes; K1 and K2 require only $d$ divisible by $32$.

\fig{latency} reports kernel latency and effective HBM bandwidth as a function of token count $m$. K2 is ${\sim}1.4\times$ faster than FWHT throughout and reaches $85$--$88\%$ of Scale's bandwidth while applying a full-rank $32 \times 32$ tensor-core GEMM per block: the block-diagonal structure confines each tile to one coalesced read and one coalesced write with no inter-block communication, whereas FWHT's global coupling forces multiple passes per output tile.

\begin{table}[t]
  \caption{Kernel latency ($\mu$s) at power-of-two $d$ dimensions, MI350, FP16. All methods fuse the MXFP4e2 output quantizer; \textbf{Q4} reports the bare quantizer kernel as a lower bound. K2 uses the \emph{best} of the on-the-fly variants (rematerialization vs.\ data-movement) per shape. Speedup ratios are reported as "Method A / Method B" meaning "Method B is $X\times$ faster than Method A"}
  \lbltbl{layer_shapes}
  \centering
  \resizebox{\textwidth}{!}{%
  \begin{tabular}{rrrrrrrrrr}
    \toprule
    $m$ & $d$ & \shortstack[r]{Q4 \\($\mu$s)} & \shortstack[r]{Scale \\ ($\mu$s)} & \shortstack[r]{K1 \\ ($\mu$s)} & \shortstack[r]{K2 \\ ($\mu$s)} & \shortstack[r]{FWHT \\ ($\mu$s)} & \shortstack[r]{Scale/K1\\Speedup} & \shortstack[r]{Scale/K2\\Speedup} & \shortstack[r]{FWHT/K2\\Speedup} \\
    \midrule
    4096 &  1024 &  9.40 & 10.56 &  9.96 & 14.00 &  18.16 & $1.06\times$ & $0.75\times$ & $1.30\times$ \\
    4096 &  4096 & 22.64 & 25.52 & 23.88 & 27.80 &  68.04 & $1.07\times$ & $0.92\times$ & $2.45\times$ \\
    4096 & 16384 & 74.44 & 75.44 & 75.32 & 78.52 & 186.04 & $1.00\times$ & $0.96\times$ & $2.37\times$ \\
     512 &  4096 &  7.20 &  9.24 &  8.12 & 10.28 &  14.40 & $1.14\times$ & $0.90\times$ & $1.40\times$ \\
     512 & 16384 & 13.68 & 16.36 & 15.36 & 19.32 &  33.96 & $1.06\times$ & $0.85\times$ & $1.76\times$ \\
    \bottomrule
  \end{tabular}}
\vspace{-10pt}
\end{table}

K1 is at least as fast as per-channel scaling on every tested shape ($1.00$--$1.14\times$): a single $32\!\times\!32$ tensor-core GEMM per block extracts more useful throughput than scaling's scalar elementwise multiplies, even though the diagonal operation is cheaper on paper. K2 loses to Scale by at most $25\%$ ($0.75$--$0.96\times$): rematerializing the $32\!\times\!32$ Kronecker block from its $15$ parameters adds register pressure inside each workgroup, and that materialization is redundant across the workgroups that operate on the same column tile, so the savings in HBM traffic are partially offset by per-thread arithmetic. Both variants nonetheless deliver consistent acceleration over FWHT, with K2 reaching up to $2.45\times$ on larger matrices. The Q4 column shows that at larger $d$ the fused quantizer already dominates kernel time, leaving little headroom between any transform variant and the bare quantizer. Unlike FWHT, K1 and K2 handle arbitrary dimensions natively, avoiding the padding or truncation that non-power-of-two layers would require in production.

\tbl{layer_shapes} reports latency on power-of-two $d$ representative of FLUX.1 and PixArt-$\Sigma$ shapes. K1 matches or exceeds Scale on every shape ($1.00$--$1.14\times$): a single tensor-core $32\!\times\!32$ GEMM per block extracts more throughput than Scale's scalar multiplies. K2 stays within $25\%$ of Scale ($0.75$--$0.96\times$). K1 is ${\sim}6$--$10\%$ faster than K2, reflecting the cost of on-the-fly Kronecker construction. Register pressure from the on-the-fly Kronecker reconstruction offsets part of the HBM-traffic savings. K2 compensates with $68\times$ parameter compression (15 scalars vs.\ 1024 values per block), making it the preferred variant for deployment. Both K1 and K2 deliver $1.30$--$2.45\times$ speedups over FWHT and operate on arbitrary $d$. At larger $d$ the fused Q4 quantizer dominates kernel time, leaving little headroom between any transform variant and the bare quantizer.

\vspace{-10pt}
\section{Conclusion and Limitations}\lblsect{conclusion}
\vspace{-10pt}
We presented KroQuant, a post-training quantization method for diffusion transformers that introduces learned Kronecker-structured block transforms as an online activation preprocessing step. The key observation is that AdaLN prevents the offline weight-absorption used by LLM rotation methods, and that a full-dimensional online transform is too costly. A block-diagonal learned transform parameterized via Kronecker products of $2 \times 2$ factors fills this gap at $O(d)$ parameter cost and tensor-core-native compute cost.

The ablation confirms that the learned Kronecker transform provides strictly more expressive power than a fixed Hadamard initialization: starting from Hadamard (which matches per-channel smoothing on FID), gradient-based optimization of 15 parameters per block reduces LPIPS by 0.035 and improves PSNR by 0.8~dB on PixArt-$\Sigma$. The hardware analysis shows that the transform's block-diagonal structure keeps it single-pass and memory-bound, achieving near-bandwidth-optimal execution at $1.4$--$1.9\times$ the throughput of FWHT.

KroQuant is complementary to weight-side corrections: the pipeline applies LoRaQ offline after the activation transforms are fixed, and the combination outperforms either component alone. On all three evaluated DiT architectures (PixArt-$\Sigma$, SANA, FLUX.1-schnell), KroQuant achieves the best or competitive LPIPS and PSNR at MXFP4 E2M1 W4A4.
\vspace{-5pt}
\paragraph{Limitations.} The Kronecker construction has only $3 \log_2 n$ parameters per $n \times n$ block, growing too slowly with block size to capture the full expressivity of $\mathrm{GL}(n)$ at larger blocks; KroQuant therefore remains tied to $n=32$ in practice. Joint multi-layer optimization is unstable in our experiments, partly due to the unit-determinant LU parametrization, leaving other parametrizations as future work. Finally, KroQuant outputs tend to have higher contrast than the FP16 reference, an empirical effect whose theoretical interpretation lies beyond the scope of this paper.
\vspace{-5pt}
\paragraph{Broader impacts.} A broader-impacts discussion is provided in~\app{broader_impacts}.

% TODO: add acknowledgments in camera-ready version only
% \begin{ack}
% \end{ack}

\bibliographystyle{abbrvnat}
\bibliography{neurips_2026}
\newpage
\appendix
\section*{Appendix}
\section{Kernel performance on production layer shapes}\lblapp{appendix-kernel}

\tbl{layer_shapes_actual} reports kernel latency on production linear-layer shapes drawn from FLUX.1 and PixArt-$\Sigma$. As in~\tbl{layer_shapes}, every variant fuses the MXFP4e2 output quantizer; \textbf{Q4} is the bare fused-quantizer kernel and serves as a lower bound, and \textbf{K2} reports the best of the on-the-fly variants (rematerialization vs.\ data-movement) per shape. FWHT is omitted because most of these shapes are not power-of-two; the fair comparison including FWHT at padded power-of-two dimensions is~\tbl{layer_shapes}.

\begin{table}[hb]
  \caption{Kernel latency ($\mu$s) on production layer shapes from FLUX.1 and PixArt-$\Sigma$, MI350, FP16. All methods fuse the MXFP4e2 output quantizer. Speedup ratios are reported as Method A / Method B, meaning "Method B is $X\times$ faster than Method A".}
  \lbltbl{layer_shapes_actual}
  \centering
  \begin{tabular}{rrrrrrrr}
    \toprule
    $m$ & $d$ & \shortstack[r]{Q4 \\ ($\mu$s)} & \shortstack[r]{Scale \\ ($\mu$s)} & \shortstack[r]{K1 \\ ($\mu$s)} & \shortstack[r]{K2 \\ ($\mu$s)} & \shortstack[r]{Scale/K1\\Speedup} & \shortstack[r]{Scale/K2\\Speedup} \\
    \midrule
     512 &  3072 &  6.72 &  9.04 &  7.64 &  8.84 & $1.18\times$ & $1.02\times$ \\
     512 & 12288 & 11.80 & 14.68 & 13.08 & 17.00 & $1.12\times$ & $0.86\times$ \\
    4096 &  1152 & 10.64 & 14.00 & 10.68 & 14.64 & $1.31\times$ & $0.96\times$ \\
    4096 &  3072 & 18.56 & 21.12 & 19.28 & 23.32 & $1.10\times$ & $0.91\times$ \\
    4096 &  3456 & 20.60 & 24.20 & 21.32 & 25.36 & $1.14\times$ & $0.95\times$ \\
    4096 &  4096 & 22.08 & 25.80 & 23.36 & 27.28 & $1.10\times$ & $0.95\times$ \\
    4096 &  4608 & 25.72 & 28.40 & 26.40 & 30.36 & $1.08\times$ & $0.94\times$ \\
    4096 & 12288 & 59.64 & 59.12 & 59.92 & 62.80 & $0.99\times$ & $0.94\times$ \\
    \bottomrule
  \end{tabular}
\end{table}

K1 matches or improves on per-channel scaling on every shape except the largest (Scale/K1 ranges from $0.99\times$ at $d{=}12288$ to $1.31\times$ at $d{=}1152$), confirming on production dimensions the same pattern as~\tbl{layer_shapes}: a fused $32{\times}32$ tensor-core GEMM per block beats Scale's scalar elementwise multiplies on bandwidth-limited shapes. K2 stays within roughly $7\%$ of Scale on most shapes (Scale/K2 between $0.91\times$ and $1.02\times$); the worst case is $(m{=}512, d{=}12288)$ at $0.86\times$, where the small-batch regime exposes the rematerialization cost of K2's on-the-fly Kronecker reconstruction against a Scale kernel whose tile is fully utilized. At $(m{=}4096, d{=}12288)$ the bare Q4 quantizer alone takes $59.64\,\mu$s, leaving negligible headroom above this lower bound for any transform variant. K1 and K2 handle these arbitrary $d$ values natively, avoiding the padding or truncation that FWHT would require on non-power-of-two layers.

\section{K2 streaming-product kernel}\lblapp{k2-kernel}

K2 evaluates the Kronecker formula of \sect{method-5} as a running product over the five factors $G_1, \dots, G_5$. Each step extracts one bit of the row index and one bit of the column index, looks up the corresponding entry of the current $G_k$ via a vectorized scalar broadcast, and multiplies the running tile by the result element-wise.

\begin{verbatim}
i = tl.arange(0, 32)[:, None]   # row indices, shape (32, 1)
j = tl.arange(0, 32)[None, :]   # col indices, shape (1, 32)

# Step 1: K = G1[bit_4(r), bit_4(c)]
ir = (i >> 4) & 1; jc = (j >> 4) & 1
K = tl.where(ir == 0,
             tl.where(jc == 0, g1_00, g1_01),
             tl.where(jc == 0, g1_10, g1_11))

# Steps 2-5: K *= G_k[bit_{5-k}(r), bit_{5-k}(c)], k = 2, 3, 4, 5
for shift, (g00, g01, g10, g11) in zip(
        (3, 2, 1, 0),
        ((g2_00, g2_01, g2_10, g2_11),
         (g3_00, g3_01, g3_10, g3_11),
         (g4_00, g4_01, g4_10, g4_11),
         (g5_00, g5_01, g5_10, g5_11))):
    ir = (i >> shift) & 1; jc = (j >> shift) & 1
    K = K * tl.where(ir == 0,
                     tl.where(jc == 0, g00, g01),
                     tl.where(jc == 0, g10, g11))
\end{verbatim}

Each \verb|tl.where(ir==0, tl.where(jc==0, g_00, g_01), tl.where(jc==0, g_10, g_11))| broadcasts the four scalar entries of $G_k$ across the $(32, 32)$ tile, selecting the correct one at every position from a single bit of the row and a single bit of the column. The element-wise multiplication accumulates the product $\prod_k G_k[r_{5-k}, c_{5-k}]$ that defines $K[r,c]$.

At each step only two $(32{\times}32)$ register tiles are live: the running product $K$ and the temporary factor tile produced by \verb|tl.where|. The temporary is consumed immediately and freed before the next step. With \verb|BLOCK_ROWS=128| and 8 warps (256 threads), each $(32, 32)$ tile occupies 4 registers per thread, so the kernel pays at most 8 registers for the Kronecker construction itself. That budget is what leaves headroom to fuse the downstream MXFP4e2 quantizer (which requires its own bit-manipulation intermediates) into the same kernel.

\section{Normalization--rotation commutativity}\lblapp{commute}

For any invertible linear transform $T \in \mathrm{GL}(d)$ applied to a normalization's input to be absorbed offline into adjacent weight matrices, the normalization must satisfy $\mathrm{Norm}(xT)\,T^{-1} = \mathrm{Norm}(x)$. We test the orthogonal case $T = R \in O(d)$ as the easiest instance: any failure here generalizes to non-orthogonal but invertible $T$. ~\tbl{commute} reports the maximum absolute deviation $\big\|\mathrm{Norm}(XR)R^\top - \mathrm{Norm}(X)\big\|_\infty$ for a random orthogonal $R \in \mathbb{R}^{64\times 64}$ on a random batch $X \in \mathbb{R}^{32\times 64}$ (seed 42). RMSNorm matches floating-point precision because its denominator depends only on $\|x\|_2$, which orthogonal rotations preserve; it would not commute with a non-orthogonal $T$. LayerNorm, BatchNorm, GroupNorm, AdaLN, and AdaRMSNorm all incur $O(1)$ deviation: the normalization either subtracts an axis-aligned mean, or applies a learned per-channel scale/shift, or both (AdaLN). Because both failure modes are axis-aligned, they cannot invert any cross-channel transform applied to the normalization's input; only per-channel input scaling can fold into the affine.

\begin{table}[h]
  \caption{Maximum absolute deviation between $\mathrm{Norm}(XR)R^\top$ and $\mathrm{Norm}(X)$ for common normalizations under a random orthogonal rotation $R$. Smaller is better; only RMSNorm commutes, and only with orthogonal $R$.}
  \lbltbl{commute}
  \centering
  \begin{tabular}{lcc}
    \toprule
    Normalization & Max abs.\ error & Commutes with $R \in O(d)$? \\
    \midrule
    RMSNorm     & $1.7 \times 10^{-6}$ & \cmark \\
    LayerNorm   & $9.2 \times 10^{-1}$ & \xmark \\
    BatchNorm   & $1.3$                & \xmark \\
    GroupNorm   & $2.1$                & \xmark \\
    AdaLN       & $4.2$                & \xmark \\
    AdaRMSNorm  & $4.4$                & \xmark \\
    \bottomrule
  \end{tabular}
\end{table}

\section{Implementation Details}\lblapp{config}

\paragraph{Quantization.}~The activations forwarded to the low-rank branch are kept at 16-bit (full-precision). FP16 is used for Pixart-$\Sigma$ while SANA and Flux.1 leverage BF16 by default. The activations forwarded to the residual branch and the residual weights are quantized to MXFP4e2 and a W4A4 matrix multiplication is applied. SVDQuant keeps the low-rank branch at full-precision for a rank of 32. LoRaQ and Kroquant by extension quantizes the low-rank branch in MXFP4e2 with a rank of 128 for an equivalent memory overhead than SVDQuant. All quantization operators (MXFP4e2, the FP8 E8M0 block scale, and the underlying MX block formatting) are implemented via the TensorCast library~\citep{tensorcast}.

\subsection{KroQuant}
\paragraph{Calibration.}~For each linear layer we optimize the Kronecker-factored block transform $T$ by minimizing the objective~\eqn{objective} using Adam for 100 steps on the 128 calibration activations with a learning rate of $10^{-2}$. The calibration samples are the same prompts than~\citet{li2025svdquant, loraq}. 128 activations are randomly selected for each layer following the same seed configuration as~\citet{li2025svdquant}. Those activations are kept and used across all experiments for fair comparison.

\subsection{Smoothing}
\paragraph{Smoothing Calibration}~Considering that our method aims to replace this calibration used by our baselines. As a first processing calibration,~\citet{li2025svdquant,loraq} apply per-channel smoothing~\citep{xiao2023smoothquant} with the layer-wise tuning of~\citet{li2025svdquant}: for each linear layer, a vector $\gamma \in \sR^d$ is built component-wise as
$$\gamma_i = \frac{(\max_j \abs{\mX_{j,i}})^\alpha}{(\max_j \abs{\mW_{i,j}})^\beta}, \qquad i \in \{0, \ldots, d-1\},$$
with the migration strengths $\alpha, \beta$ selected layer by layer to minimize the output mean-squared error of a downstream module that consumes the quantized layer's output. During this search, the weights are quantized through a rank-$32$ FP16 SVD truncation, and the search uses the same calibration dataset as KroQuant (\sect{exp-1}).

\subsection{LoRaQ}
After Kroquant or Smoothing is fixed, offline LoRaQ weight calibration (\sect{method-4}) is applied with rank $r = 128$ in MXFP4e2 low rank branches for every layer.
\paragraph{Optimization.} We optimize LoRaQ's objective for any experiment implying it with Adam~\citep{kingma2014adam} using a learning rate of $10^{-3}$ and 500 steps per weight. The quantization-aware fine tuning of the low-rank components using a rotation transformation is done with a Cayley SGD algorithm at a learning rate of $5 \cdot 10^{-1}$ and 500 steps per weight, following~\citet{loraq}.

\subsection{SVDQuant}
\paragraph{Calibration.}~After the smoothing calibration,~\citet{li2025svdquant} calibrate their low-rank components using an iterative algorithm which we follow as well to reproduce the method in MXFP4e2: 100 candidates are recursively computed. An MSE loss is computed at the output level of the linear layer or transformer block, depending on the linear layer to calibrate. Each candidate is evaluated in the recursive order. A stopping algorithm returns the first encountered minimum local in the objective space with the associated low rank candidate which is then used as the low rank components for the auxiliary branch.

\paragraph{Compute.} Calibration of all linear layers of one DiT (PixArt-$\Sigma$, SANA) runs on a single NVIDIA H100 within a few hours of wall-clock per model. Due to its size Flux.1 required a NVIDIA H200 for calibration. In practice we used 4 GPUs at a time for Smoothing, Kroquant and generating images. Kernel-latency benchmarks (\sect{exp-4}) run on an AMD Instinct MI350 in FP16 mode, with kernels implemented in Triton~3.6.0 against ROCm~7.2.2.

\paragraph{Asset licenses.} We use the following external assets under their respective licenses: \textbf{PixArt-$\Sigma$}~\citep{chen2024pixart-Sigma} (CreativeML Open RAIL++-M); \textbf{SANA}~\citep{xie2025sana} (Apache 2.0); \textbf{FLUX.1-schnell}~\citep{flux1} (Apache 2.0); \textbf{MJHQ-30K}~\citep{li2024mjhq}; \textbf{sDCI}~\citep{urbanek2024sdci}; \textbf{COCO 2017}~\citep{chen2015coco} (CC BY 4.0 for annotations, with underlying images subject to their original Flickr licenses); and \textbf{TensorCast}~\citep{tensorcast} (MIT). All assets are used in accordance with their applicable license terms and restrictions.
\section{Broader Impacts}
\lblapp{broader_impacts}

KroQuant accelerates inference of three publicly released diffusion transformers (PixArt-$\Sigma$, SANA, FLUX.1-schnell) without altering their generative capabilities, training data, or safety tuning. We discuss both directions of its societal impact and the boundary of what changes relative to a baseline FP16 deployment of the same upstream checkpoints.

\paragraph{Positive consequences.}~A method that delivers near-FP16 image quality at W4A4 with tensor-core-native online transforms reduces the per-inference energy and memory cost of large text-to-image models. The resulting deployment cost reduction broadens access to current state-of-the-art image generation: lower hardware tiers become viable for academic research, classroom use, and applied work in resource-constrained settings, and large-scale services see a smaller carbon footprint per generated sample. By keeping the rotation parameter count at $O(d)$ and avoiding any retraining, KroQuant also lowers the engineering and compute barrier to deploying public DiTs on commodity accelerators.

\paragraph{Negative consequences.}~The same efficiency improvement makes the dual-use risks of upstream diffusion transformers cheaper to operate at scale. Non-consensual deepfakes, targeted disinformation, and copyright-infringing synthesis become accessible to actors that previously could not afford the inference cost, and a sub-16-bit pipeline that runs on a single commodity accelerator widens the population of potential misuse operators relative to an FP16 deployment.

\paragraph{Boundary of incremental impact.}~KroQuant introduces no new generative capability, no new modality, no relaxation of safety tuning, and no new dataset or pretrained-model release at submission. The qualitative threat model of the upstream checkpoints (PixArt-$\Sigma$ under OpenRAIL-M, SANA and FLUX.1-schnell under Apache 2.0) is therefore unchanged; only the cost curve of operating those models shifts. Responsibility for misuse mitigation remains with the original upstream license terms, which already bind the released weights and are not relaxed by quantization. We treat KroQuant's incremental societal impact as bounded by that of the upstream models it accelerates, while acknowledging that any inference-cost reduction is amplified at deployment scale.

\section{Additional Visual Results}\lblapp{vis}
\begin{figure*}[htbp]
    \centering
    \caption{Comparison of images generated by PixArt-$\Sigma$ across different quantization configurations following~\tbl{main} settings.}
    \lblfig{mxfp4_appendix_comparison}
    %\vspace{-2pt}
    \setlength{\tabcolsep}{3pt} % Adjust column separation
    \begin{tabular}{c@{\hspace{1em}}c@{\hspace{1em}}c@{\hspace{1em}}c@{\hspace{1em}}}
        \toprule
        \textbf{FP16} & \textbf{SVDQuant} & \textbf{LoRaQ} & \textbf{KroQuant} \\
        \midrule
        
        % --- ROW 1 ---
        \adjustbox{frame,width=0.18\textwidth}{\includegraphics{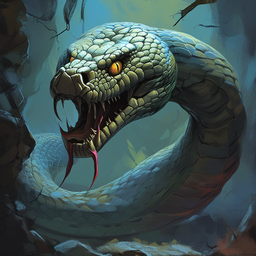}} &
        \adjustbox{frame,width=0.18\textwidth}{\includegraphics{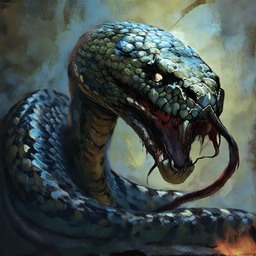}} &
        \adjustbox{frame,width=0.18\textwidth}{\includegraphics{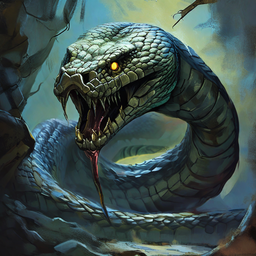}} &
        \adjustbox{frame,width=0.18\textwidth}{\includegraphics{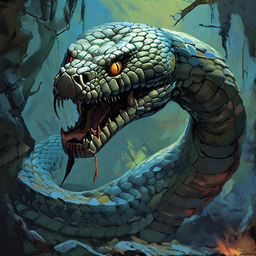}} \\
        
        % --- ROW 2 ---
        \adjustbox{frame,width=0.18\textwidth}{\includegraphics{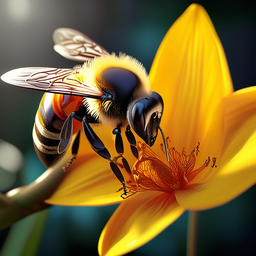}} &
        \adjustbox{frame,width=0.18\textwidth}{\includegraphics{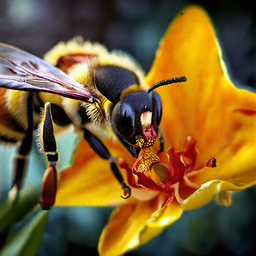}} &
        \adjustbox{frame,width=0.18\textwidth}{\includegraphics{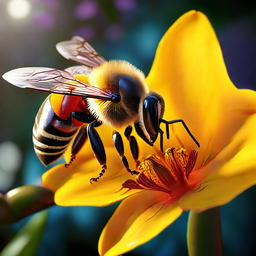}} &
        \adjustbox{frame,width=0.18\textwidth}{\includegraphics{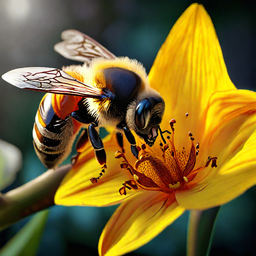}} \\

        % --- ROW 3 ---
        \adjustbox{frame,width=0.18\textwidth}{\includegraphics{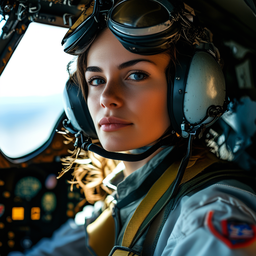}} &       
        \adjustbox{frame,width=0.18\textwidth}{\includegraphics{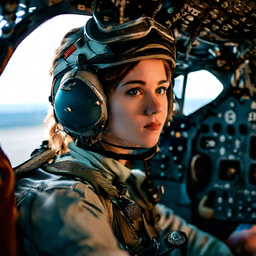}} &        
        \adjustbox{frame,width=0.18\textwidth}{\includegraphics{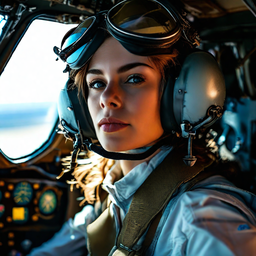}} &         
        \adjustbox{frame,width=0.18\textwidth}{\includegraphics{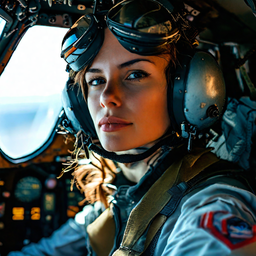}} \\

        % --- ROW 4 ---
        \adjustbox{frame,width=0.18\textwidth}{\includegraphics{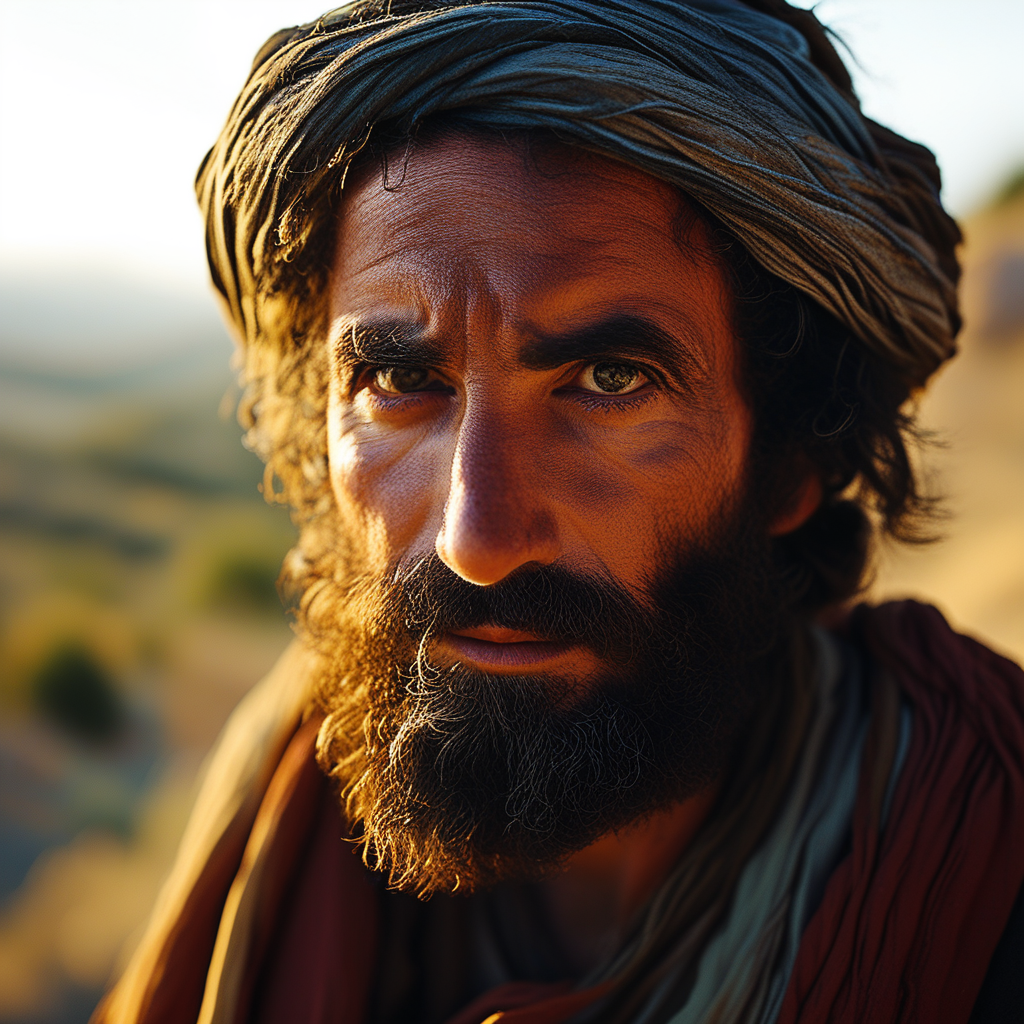}} &
        \adjustbox{frame,width=0.18\textwidth}{\includegraphics{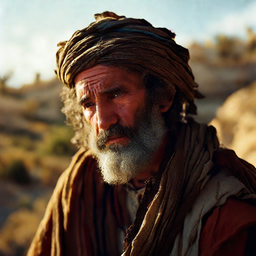}} &
        \adjustbox{frame,width=0.18\textwidth}{\includegraphics{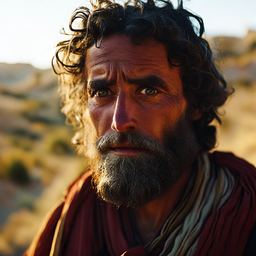}} &
        \adjustbox{frame,width=0.18\textwidth}{\includegraphics{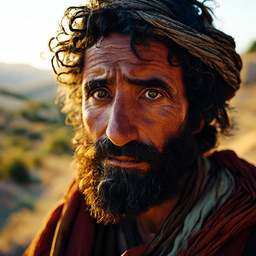}} \\

        % --- ROW 5 ---
        \adjustbox{frame,width=0.18\textwidth}{\includegraphics{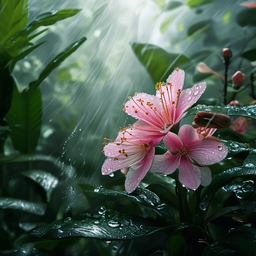}} &
        \adjustbox{frame,width=0.18\textwidth}{\includegraphics{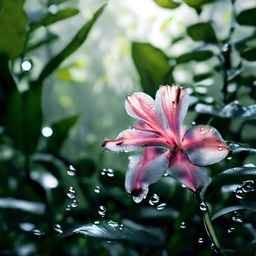}} &
        \adjustbox{frame,width=0.18\textwidth}{\includegraphics{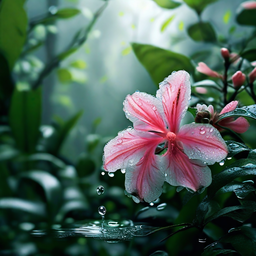}} &
        \adjustbox{frame,width=0.18\textwidth}{\includegraphics{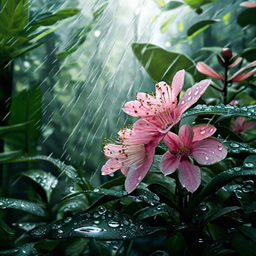}} \\

        % --- ROW 6 ---
        \adjustbox{frame,width=0.18\textwidth}{\includegraphics{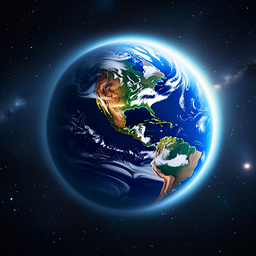}} &
        \adjustbox{frame,width=0.18\textwidth}{\includegraphics{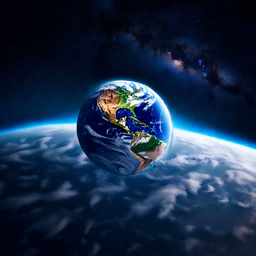}} &
        \adjustbox{frame,width=0.18\textwidth}{\includegraphics{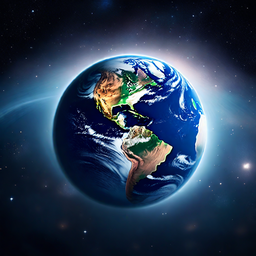}} &
        \adjustbox{frame,width=0.18\textwidth}{\includegraphics{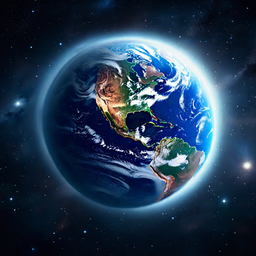}} \\

        % --- ROW 7 ---
        \adjustbox{frame,width=0.18\textwidth}{\includegraphics{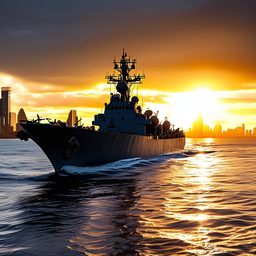}} &
        \adjustbox{frame,width=0.18\textwidth}{\includegraphics{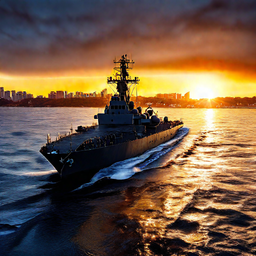}} &
        \adjustbox{frame,width=0.18\textwidth}{\includegraphics{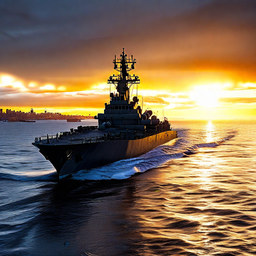}} &
        \adjustbox{frame,width=0.18\textwidth}{\includegraphics{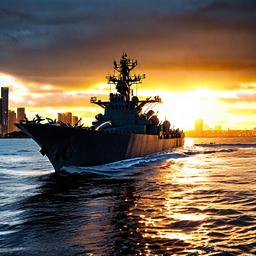}} \\

        % --- ROW 8 ---
        \adjustbox{frame,width=0.18\textwidth}{\includegraphics{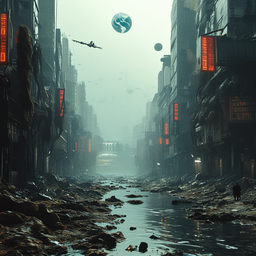}} &
        \adjustbox{frame,width=0.18\textwidth}{\includegraphics{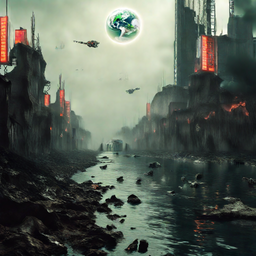}} &
        \adjustbox{frame,width=0.18\textwidth}{\includegraphics{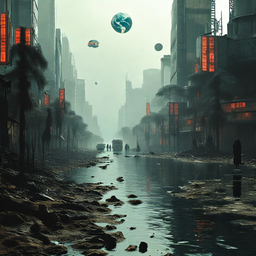}} &
        \adjustbox{frame,width=0.18\textwidth}{\includegraphics{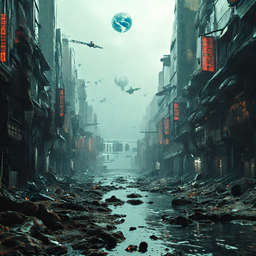}} \\
    \end{tabular}
\end{figure*}
% TODO: additional appendix sections (additional ablations)

% Checklist removed for arXiv preprint (submission-only artifact)
% \newpage
% \input{checklist}

\end{document}